\documentclass[letterpaper, 10 pt, conference]{ieeeconf}  

\IEEEoverridecommandlockouts                              

\usepackage{amsmath} 
\usepackage{amssymb}  

\usepackage{graphicx}
\usepackage{caption}
\usepackage{booktabs}
\usepackage[linesnumbered,ruled,vlined]{algorithm2e}
\usepackage{float}
\usepackage{cite}
\usepackage{color}
\usepackage[table]{xcolor}
\usepackage{multirow}
\usepackage{fontawesome5}
\usepackage{hyperref}
\usepackage{fancyhdr}
\usepackage{xspace}
\usepackage{pifont}
\usepackage{makecell}
\usepackage{pifont}

\hypersetup{colorlinks=true, linkcolor=blue!50!black,
            citecolor=blue!50!black, urlcolor=blue!50!black}

\title{\textbf{Verti-WM: A Physics-Aided Exteroceptive World Model for \\ Off-Road Reinforcement Learning
}}

\author{Chenhui Pan$^{*}$, Tong Xu$^{*}$, and Xuesu Xiao
\thanks{All authors are with the Department of Computer Science, George Mason University {\tt\small \{cpan7, txu25, xiao\}@gmu.edu}}
\thanks{*Equally contributing authors}
}

\begin{document}

\maketitle
\thispagestyle{empty}
\pagestyle{empty}

\begin{abstract}
Reinforcement learning for off-road navigation requires extensive vehicle-terrain interaction data, which are costly to collect in high-fidelity simulation. World models offer a promising alternative by replacing simulator rollouts during policy optimization. However, an off-road world model must condition state transitions on exteroceptive terrain information, which proprioception alone does not provide. This challenge is further amplified by the need to model both rigid and deformable terrain, where data-driven and physics-based approaches offer complementary strengths. We propose \textit{Verti-WM}, a physics-aided exteroceptive world model that recurrently fuses a frozen Transformer for rigid terrain and a neuro-symbolic terramechanics model for deformable terrain. Elevation and semantic observations queried from a supplied map at each predicted pose condition fusion, enabling six-degree-of-freedom rollouts for policy optimization without further simulator access. Verti-WM reduces prediction error by 34.6\% and 21.7\% over data-driven and physics-based baselines, respectively. Policies trained entirely within Verti-WM achieve comparable task success rates while reducing computation time by 23.6X relative to direct training in the high-fidelity simulator. We further validate Verti-WM using real-world data, enabling policy optimization within learned real-world kinodynamics and achieving a 80\% success rate on the Verti-4-Wheeler platform, compared with 40\% for direct sim-to-real transfer.
\end{abstract}

\section{Introduction}
\label{sec:introduction}

Off-road navigation requires vehicles to traverse terrain with varying geometry and physical properties, causing identical control inputs to produce different slip, sinkage, and six Degree-of-Freedom (DoF) motion~\cite{wong2010terramechanics,triest2022tartandrive,lee2023terrain}. Reinforcement Learning (RL) provides a framework for learning policies under these coupled vehicle-terrain interactions, but requires extensive interaction data~\cite{vertibench2025}. Collecting such data on physical vehicles is time-consuming and difficult to scale, while high-fidelity simulation is computationally expensive because vehicle-terrain interactions and contact dynamics must be repeatedly resolved~\cite{tasora2016chrono,vertibench2025}. Direct policy optimization is therefore costly in both real-world and simulation settings.

\begin{figure}[t!]
    \centering
    \includegraphics[width=\columnwidth]{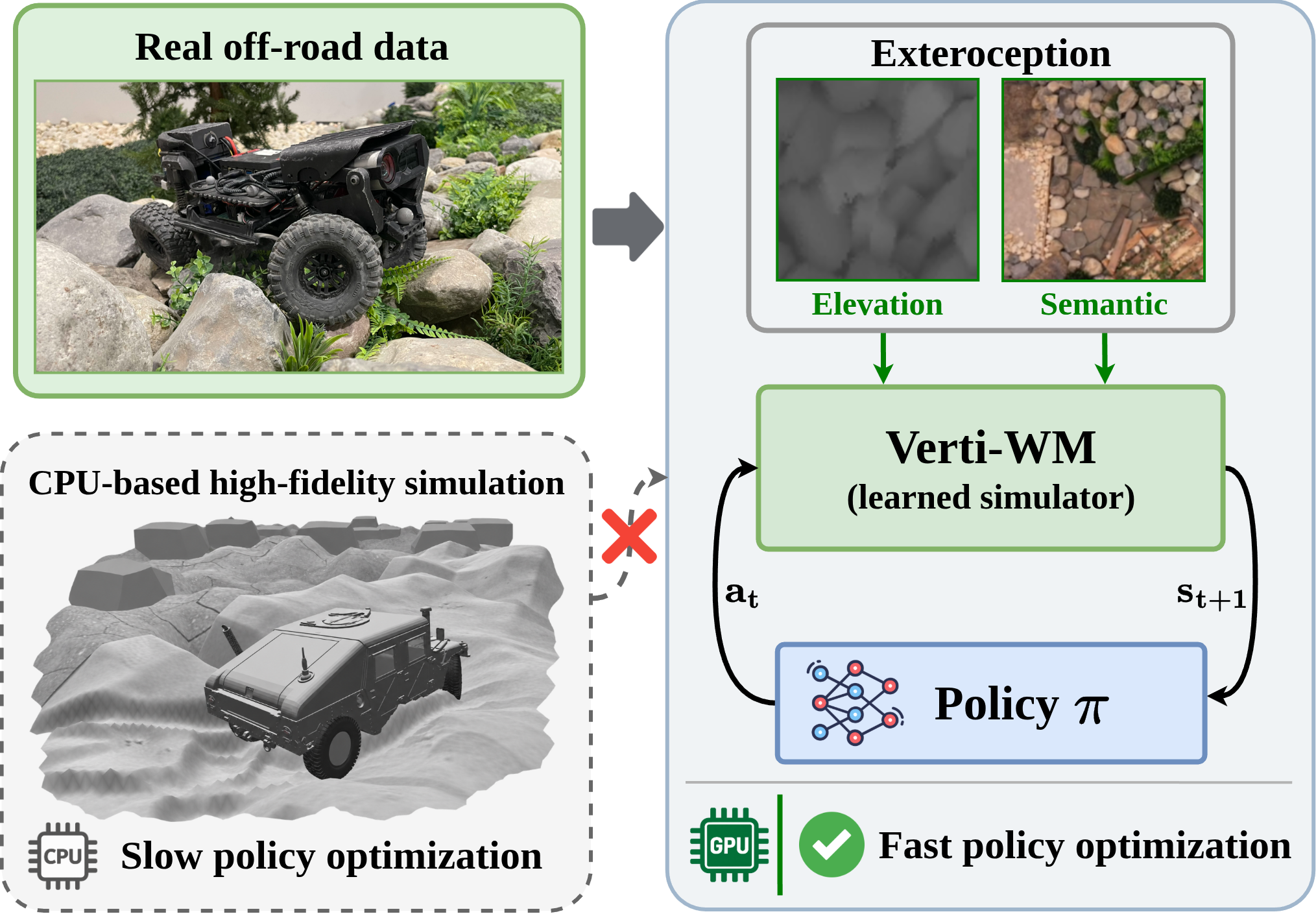}
    \caption{High-fidelity CPU-based simulation makes direct policy optimization computationally expensive. Verti-WM replaces it with a GPU-accelerated exteroceptive learned simulator trained from real off-road data, enabling faster policy optimization.}
    \label{fig::motivation}
    \vspace{-10pt}
\end{figure}

World models can reduce this cost by replacing simulator interactions during policy optimization with rollouts from learned kinodynamics~\cite{ha2018worldmodels,hafner2023dreamerv3,li2024think2drive}. Learned robotic simulators such as the Robotic World Model~\cite{li2025rwm} and Neural Robot Dynamics~\cite{xu2025nerd} further show that policies can be optimized entirely within learned environments. However, these models primarily rely on robot-centric states, actions, or contact information and do not explicitly condition vehicle motion on the local terrain encountered during traversal. Proprioception alone cannot distinguish a slope, rigid obstacle, or deformable region before its effect appears in the vehicle state. A simulator learned for off-road kinodynamics must therefore incorporate exteroceptive terrain observations to anticipate terrain-dependent motion.

An exteroceptive world model for off-road dynamics must also capture vehicle motion across both rigid and deformable terrain, where similar surface geometry can induce distinct vehicle-terrain kinodynamics. End-to-end learned kinodynamic models can effectively capture rigid-terrain behavior in a data-driven manner~\cite{lee2023terrain}, while terramechanics provides physical structure for deformation-dependent effects such as sinkage and soil shear~\cite{wong2010terramechanics,nesam}. However, purely data-driven models lack explicit structure for complex deformation effects, whereas physics-based models depend on terrain parameters that are difficult to obtain across diverse environments~\cite{wong2010terramechanics}. Neither approach alone is well suited to both regimes. A learned off-road simulator (Fig.~\ref{fig::motivation}) must therefore integrate their complementary strengths and handle transitions between them while mapping local elevation and semantic observations to consistent 6-DoF vehicle-motion rollouts.

To address these challenges, we propose \textit{Verti-WM}, a physics-aided exteroceptive world model for data-efficient off-road RL. Verti-WM combines two frozen kinodynamic specialists: a Transformer-based model for rigid terrain and a neuro-symbolic terramechanics model for deformable terrain. A recurrent gate conditioned on the vehicle state, action, elevation and semantic features, and both specialist predictions estimates separate fusion coefficients for the six body-frame pose increments. During rollout, Verti-WM queries local elevation and semantic patches from a supplied terrain map at each predicted pose rather than predicting future exteroceptive observations. Overlapping consecutive patches preserve terrain continuity and reduce error accumulation over long-horizon rollouts. This map-conditioned autoregressive formulation enables terrain-dependent 6-DoF rollouts across rigid terrain, deformable terrain, and their transitions without further access to high-fidelity simulators during policy optimization.

To our knowledge, Verti-WM is the first physics-aided exteroceptive world model for policy optimization across both rigid and deformable terrain. Policies are optimized entirely within Verti-WM and evaluated in a high-fidelity simulator, Verti-Bench~\cite{vertibench2025}. For physical validation, we construct a separate Verti-WM from real-world trajectories, optimize a policy within the learned kinodynamics, and compare its deployment performance against direct sim-to-real transfer without fine-tuning.
The main contributions of this work are:
\begin{itemize}
    \item We develop a physics-aided exteroceptive world model that fuses frozen data-driven and physics-based kinodynamic specialists through recurrent gating with separate weights for body-frame pose increments, enabling map-conditioned 6-DoF rollouts across rigid and deformable terrain.
    
    \item We show that Verti-WM reduces prediction error by 34.6\% and 21.7\% over data-driven and physics-based baselines, respectively, while policies optimized within Verti-WM achieve comparable success in Verti-Bench with 23.6X faster policy training than direct optimization in the high-fidelity simulator.
    
    \item We validate Verti-WM using real-world data, where a policy optimized within the learned kinodynamics achieves an 80\% success rate on the Verti-4-Wheeler robot~\cite{datar2024toward}, compared with 40\% for direct sim-to-real transfer.
\end{itemize}

\section{Related Work}
\label{sec:related_work}
In this section, we review related work on terrain-conditioned off-road kinodynamics, world models for navigation planning, and learned simulators for policy optimization.

\subsection{Terrain-Conditioned Off-Road Kinodynamics}

Classical terramechanics~\cite{wong2010terramechanics} provides interpretable pressure-sinkage and shear models for deformable terrain, but its accuracy depends on terrain parameters that are difficult to obtain across heterogeneous environments. Large-scale multimodal datasets, including TartanDrive and TartanDrive~2.0~\cite{triest2022tartandrive,sivaprakasam2024tartandrive2}, have enabled data-driven modeling of off-road vehicle-terrain kinodynamics. Recent methods incorporate exteroceptive observations into kinodynamic prediction for model-based planning through terrain-aware latent representations~\cite{lee2023terrain,datar2024terrain,xiao2021learning}, visual terrain features~\cite{gibson2024visualterrain,pokhrel2024cahsor,karnan2022vi}, and friction estimation~\cite{fu2025anynav,rabiee2019friction}. Physics-learning approaches introduce additional structure: PIAug~\cite{maheshwari2023piaug} augments training data with vehicle dynamics, while NeSAM~\cite{nesam} combines differentiable terramechanics with learned residual kinodynamics for deformable terrain. VertiAdaptor~\cite{vertiadaptor} and VertiAKD~\cite{vertiakd} further enable online adaptation of terrain-conditioned kinodynamics. These methods improve terrain-aware prediction and model-based planning, but are not designed to serve as long-horizon learned environments for policy optimization.

\subsection{World Models for Navigation Planning}

World models can support navigation by predicting future observations or latent scene representations that guide planning and action selection. Navigation World Models (NWM)~\cite{bar2025navigation} predict action-conditioned visual futures to evaluate candidate trajectories proposed by an external policy. MILE~\cite{mile2022} jointly learns latent scene dynamics and an imitation policy from expert driving videos, using imagined latent trajectories to support action prediction.

For off-road navigation, TerrainFormer~\cite{yang2026terrainformer} learns predictive bird's-eye-view terrain representations with traversability, elevation, and semantic supervision, then uses the frozen world-model features to train a decision Transformer for action prediction. Although it incorporates exteroceptive terrain information, TerrainFormer models future terrain representations rather than the vehicle motion induced by control inputs. In contrast, Verti-WM predicts explicit action-conditioned 6-DoF vehicle-terrain transitions and recursively queries local elevation and semantic observations to generate closed-loop interaction rollouts for RL policy optimization.

\subsection{World Models as Simulators for Policy Optimization}

Recent work has also explored world models as learned simulators for direct policy optimization. Think2Drive~\cite{li2024think2drive} trains an autonomous-driving policy within a compact latent world model learned from CARLA, replacing simulator interactions with imagined rollouts. Learning to Drive from a World Model~\cite{Goff2025LearningTD} similarly optimizes an on-policy driving agent in a learned visual simulator constructed from real driving data. These methods demonstrate policy optimization with perceptual world models, but focus on structured on-road environments rather than off-road vehicle-terrain interaction with terrain-dependent 6-DoF motion. Off-road RL methods instead typically rely on high-fidelity physics simulators, including learning on vertically challenging terrain~\cite{xu2024verticalrl} and TADPO~\cite{wu2026tadpo}. NRD~\cite{zhang2026nrd} is the closest learned-simulator baseline, training policies entirely within frozen reduced dynamics before evaluation in Chrono~\cite{tasora2016chrono}. However, NRD uses a binary indicator to distinguish rigid and deformable terrain types rather than local exteroceptive observations, so its state transitions do not explicitly capture terrain geometry or semantics at each vehicle pose.

\begin{figure*}[t!]
    \centering
    \includegraphics[width=\textwidth]{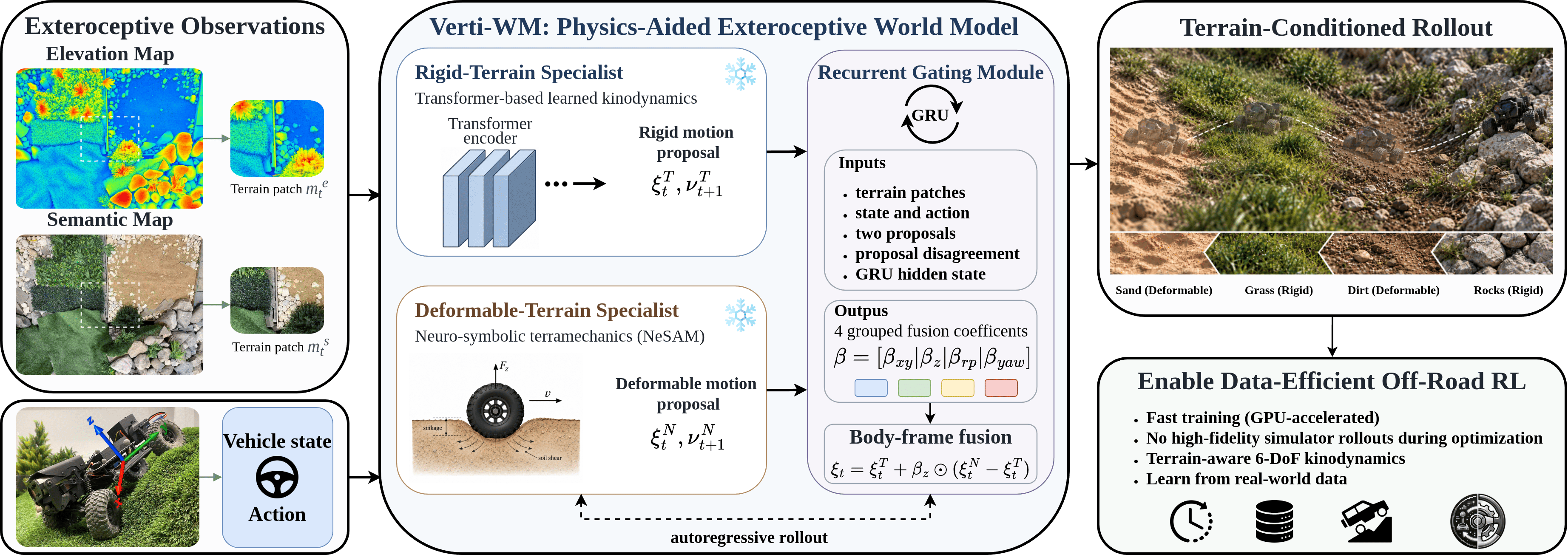}
    \caption{Overview of Verti-WM: Elevation and semantic observations queried at the predicted vehicle pose condition two frozen kinodynamic specialists. A recurrent gate fuses their body-frame motion proposals, and the resulting state determines the next terrain query. The gate is trained using multi-step trajectory prediction errors.}
    \label{fig::verti-wm_framework}
    \vspace{-10pt}
\end{figure*}

Verti-WM integrates exteroceptive terrain observations into learned-simulator policy optimization. At each predicted pose, it queries local elevation and semantic information from a supplied terrain map to predict the next 6-DoF vehicle-motion transition. A recurrent gate fuses frozen data-driven and physics-aided kinodynamic specialists, enabling terrain-dependent rollouts across rigid terrain, deformable terrain, and their transitions.

\section{Method}
\label{sec:method}

As illustrated in Fig.~\ref{fig::verti-wm_framework}, Verti-WM combines a Transformer specialist $F_T$ and a neuro-symbolic specialist $F_N$ through a recurrent gate $G_{\theta}$ with trainable parameters $\theta$. We formulate the prediction problem in Sec.~\ref{sec:method_problem}, describe the specialists in Sec.~\ref{sec:method_specialists}, and present fusion and gate training in Secs.~\ref{sec:method_fusion} and~\ref{sec:method_learning}.

\subsection{Problem Formulation}
\label{sec:method_problem}

We consider a vehicle operating at sampling interval $\Delta t$ with state
\[
    \mathbf{x}_t
    =[\mathbf{p}_t;\boldsymbol{\eta}_t;
      \mathbf{v}_t;\boldsymbol{\omega}_t]
    \in\mathbb{R}^{12},
\]
where $\mathbf{p}_t$ is world-frame position, $\boldsymbol{\eta}_t$ contains roll, pitch, and yaw, and $\mathbf{v}_t,\boldsymbol{\omega}_t$ are body-frame linear and angular velocities. Each component is three-dimensional, and $[\,;\,]$ denotes concatenation. We denote the corresponding rigid pose by $\mathbf{T}_t\in\mathrm{SE}(3)$ and collect velocities as $\boldsymbol{\nu}_t=[\mathbf{v}_t;\boldsymbol{\omega}_t]$. The control $\mathbf{u}_t\in\mathbb{R}^{2}$ contains commanded speed and steering angle. Hats distinguish fused predictions $\hat{\mathbf{x}}_t,\hat{\mathbf{T}}_t$ from reference states and poses.

The Transformer history $\mathbf{h}^{T}_t$ retains preceding state, control, and encoded terrain inputs; the neuro-symbolic history $\mathbf{h}^{N}_t$ additionally contains internal terramechanics memory. Together with the gate hidden state $\mathbf{h}^{G}_t$, they form $\mathbf{h}_t=(\mathbf{h}^{T}_t,\mathbf{h}^{N}_t,\mathbf{h}^{G}_t)$, initialized as described in Sec.~\ref{sec:method_learning}.

Given a terrain map $\mathcal{M}$, the query operator $\mathcal{Q}$ extracts vehicle-aligned elevation and semantic patches $\mathbf{m}_t=(\mathbf{m}^{e}_t,\mathbf{m}^{s}_t)$. Verti-WM defines the transition
\[
\begin{aligned}
    \mathbf{m}_t
    &=\mathcal{Q}(\mathcal{M};\hat{\mathbf{T}}_t),\\
    (\hat{\mathbf{x}}_{t+1},\mathbf{h}_{t+1})
    &=F_{\Theta}
      (\hat{\mathbf{x}}_t,\mathbf{u}_t,
       \mathbf{m}_t,\mathbf{h}_t),
\end{aligned}
\]
where $\Theta$ collects all model parameters. Repeated application generates trajectories without subsequent reference-state feedback, while terrain observations are queried rather than predicted.

The crop extent and sampling interval must ensure
\[
    \mathcal{B}(\hat{\mathbf{T}}_{t+1})
    \subseteq\Omega(\mathbf{m}_t),
\]
where $\mathcal{B}$ and $\Omega$ denote the vehicle support footprint and patch footprint in map coordinates. Thus, the current patch covers the terrain supporting the next predicted pose, over which consecutive patches overlap.

\subsection{Terrain-Conditioned Kinodynamic Specialists}
\label{sec:method_specialists}

We train $F_T$ on rigid-terrain trajectories $\mathcal{D}_r$ and $F_N$ on deformable-terrain trajectories $\mathcal{D}_d$. The Transformer predicts motion directly from interaction histories. The neuro-symbolic model follows NeSAM~\cite{nesam}, combining learned wheel-interaction variables and soil parameters with differentiable terramechanics~\cite{wong2010terramechanics}, Newton--Euler dynamics, and learned state corrections.

For specialist index $j\in\{N,T\}$, identifying the neuro-symbolic and Transformer models respectively, elevation and semantic encoders $E_j^e,E_j^s$ produce spatial feature tensors $\mathbf{Z}^{j,e}_t,\mathbf{Z}^{j,s}_t$:
\begin{equation}
\begin{aligned}
    \mathbf{Z}^{j,e}_t
    &=E^e_j(\mathbf{m}^e_t),\\
    \mathbf{Z}^{j,s}_t
    &=E^s_j(\mathbf{m}^s_t),\\
    \mathbf{z}^{j}_t
    &=[\operatorname{vec}(\mathbf{Z}^{j,e}_t);
       \operatorname{vec}(\mathbf{Z}^{j,s}_t)].
\end{aligned}
\label{eq:specialist_features}
\end{equation}
Here, $\operatorname{vec}$ flattens each tensor into the specialist's terrain representation $\mathbf{z}^{j}_t$. Both encoder pairs process the same patches but retain their respective pretrained weights. Each specialist then proposes
\begin{equation}
    \mathbf{x}^{j}_{t+1}
    =F_j(\hat{\mathbf{x}}_t,\mathbf{u}_t,
         \mathbf{z}^{j}_t,\mathbf{h}^{j}_t),
    \label{eq:specialist_proposals}
\end{equation}
where $\mathbf{x}^{j}_{t+1}$ contains candidate pose $\mathbf{T}^{j}_{t+1}$ and velocities $\boldsymbol{\nu}^{j}_{t+1}$. After independent training, both specialists, their encoders, and the soil predictor remain frozen. Both specialists are evaluated at every step while the gate is trained on mixed-terrain trajectories $\mathcal{D}_m$.

\subsection{Recurrent Kinodynamic Fusion}
\label{sec:method_fusion}

Specialist accuracy can vary across terrain conditions and motion components. To combine their predictions consistently, we express both candidate poses relative to the current body frame:
\begin{equation}
    \boldsymbol{\xi}^{j}_t
    =\operatorname{Log}
      \left(\hat{\mathbf{T}}_t^{-1}
            \mathbf{T}^{j}_{t+1}\right)^{\vee}
    \in\mathbb{R}^{6}.
    \label{eq:body_increment}
\end{equation}
The operator $\operatorname{Log}(\cdot)^{\vee}$ returns three translational tangent coordinates followed by a three-dimensional rotation vector. Their difference $\mathbf{d}_t=\boldsymbol{\xi}^{N}_t-\boldsymbol{\xi}^{T}_t$ describes the available correction to the Transformer proposal.

The gate reuses both specialists' terrain features without introducing an additional encoder. Spatial averaging within each channel, denoted by $\operatorname{Avg}$, produces a terrain vector $\mathbf{e}^{j}_t\in\mathbb{R}^{36}$ for each specialist; their concatenation forms $\mathbf{e}_t\in\mathbb{R}^{72}$. We define $\mathbf{r}_t\in\mathbb{R}^{18}$ to contain both motion proposals and their disagreement, and $\mathbf{c}_t\in\mathbb{R}^{104}$ as the complete gate input combining terrain, state, control, and proposal features:
\begin{equation}
\begin{aligned}
    \mathbf{e}^{j}_t
    &=[\operatorname{Avg}(\mathbf{Z}^{j,e}_t);
       \operatorname{Avg}(\mathbf{Z}^{j,s}_t)],\\
    \mathbf{e}_t
    &=[\mathbf{e}^{N}_t;\mathbf{e}^{T}_t],\\
    \mathbf{r}_t
    &=\operatorname{sg}
      ([\boldsymbol{\xi}^{N}_t;
        \boldsymbol{\xi}^{T}_t;\mathbf{d}_t]),\\
    \mathbf{c}_t
    &=[\mathbf{e}_t;\hat{\mathbf{x}}_t;
       \mathbf{u}_t;\mathbf{r}_t].
\end{aligned}
\label{eq:gate_context}
\end{equation}
Here, $\operatorname{sg}$ preserves forward values while blocking gradients. These inputs condition fusion on the local terrain, operating condition, and candidate vehicle responses.

A single fusion coefficient would impose the same combination on every motion component. Instead, we define $\boldsymbol{\alpha}_t=[\alpha_{t,1},\alpha_{t,2},\alpha_{t,3},\alpha_{t,4}]^\top$ for body-frame planar translation, translation along the body $z$ axis, rotation about the body $x/y$ axes, and rotation about the body $z$ axis, respectively. These groups distinguish planar, vertical, tilt-related, and heading-related motion. All coordinates are relative body-frame increments; sharing a coefficient within each pair treats its proposal as a vector and reduces independent fusion decisions.

The recurrent gate produces four unconstrained outputs $\mathbf{o}_t\in\mathbb{R}^{4}$ and updates its hidden state:
\begin{equation}
    (\mathbf{o}_t,\mathbf{h}^{G}_{t+1})
    =G_{\theta}(\mathbf{c}_t,\mathbf{h}^{G}_t).
    \label{eq:recurrent_gate}
\end{equation}
For $\ell=1,\ldots,4$, the fusion coefficient is
\begin{equation}
    \alpha_{t,\ell}
    =0.5+1.5\tanh(o_{t,\ell}).
    \label{eq:gate_coefficients}
\end{equation}
Since $\tanh$ ranges over $(-1,1)$, this mapping restricts coefficients to $(-1,2)$, with zero gate output giving equal weighting. Values in $[0,1]$ interpolate between proposals, while values outside this interval permit bounded extrapolation when interpolation cannot sufficiently correct their errors.

The four coefficients are expanded to six coordinate-wise weights:
\begin{equation}
    \boldsymbol{\beta}_t
    =[\alpha_{t,1},\alpha_{t,1},\alpha_{t,2},
      \alpha_{t,3},\alpha_{t,3},\alpha_{t,4}]^\top.
    \label{eq:weight_expansion}
\end{equation}
Repeated weights do not impose equal motion values: the $x$ and $y$ coordinates share $\alpha_{t,1}$ but combine their respective proposal components. The same applies to the two tilt-related coordinates.

With velocity disagreement $\mathbf{d}^{\nu}_t=\boldsymbol{\nu}^{N}_{t+1}-\boldsymbol{\nu}^{T}_{t+1}$, the fused transition is
\begin{equation}
\begin{aligned}
    \boldsymbol{\xi}_t
    &=\boldsymbol{\xi}^{T}_t
      +\boldsymbol{\beta}_t\odot\mathbf{d}_t,\\
    \hat{\mathbf{T}}_{t+1}
    &=\hat{\mathbf{T}}_t
      \operatorname{Exp}(\boldsymbol{\xi}_t^{\wedge}),\\
    \hat{\boldsymbol{\nu}}_{t+1}
    &=\boldsymbol{\nu}^{T}_{t+1}
      +\boldsymbol{\beta}_t\odot\mathbf{d}^{\nu}_t.
\end{aligned}
\label{eq:fused_transition}
\end{equation}
Here, $\odot$ denotes elementwise multiplication, and $\operatorname{Exp}(\cdot^\wedge)$ converts the fused tangent increment into a relative pose. The same weights combine corresponding body-frame velocity components. All-zero coefficients recover the Transformer proposal, while all-one coefficients recover the neuro-symbolic proposal. The resulting pose and velocities form $\hat{\mathbf{x}}_{t+1}$, which advances both specialist histories and determines the next terrain query.

\subsection{Autoregressive Gate Learning}
\label{sec:method_learning}

Each training sequence $\tau\in\mathcal{D}_m$ provides a map, reference states $\mathbf{x}_{0:K}$, and controls $\mathbf{u}_{0:K-1}$ for horizon $K\geq2$. We initialize $\hat{\mathbf{x}}_0=\mathbf{x}_0$ and reset all histories: attention histories start empty, the gate hidden vector starts at zero, and the neuro-symbolic model resets its physical memory. The first transition receives the recorded initial state, control, and terrain observations. Subsequent transitions use predicted states and recorded controls, with reference states used only for supervision.

At prediction step $k=1,\ldots,K$, define pose error $\boldsymbol{\delta}_k$ and component-group error $e_{k,\ell}$ as
\[
\begin{aligned}
    \boldsymbol{\delta}_k
    &=
    \begin{bmatrix}
        \hat{\mathbf{p}}_k-\mathbf{p}_k\\
        \operatorname{wrap}
        (\hat{\boldsymbol{\eta}}_k-\boldsymbol{\eta}_k)
    \end{bmatrix},\\
    e_{k,\ell}
    &=\|\mathbf{S}_{\ell}\boldsymbol{\delta}_k\|_2.
\end{aligned}
\]
The operator $\operatorname{wrap}$ maps angular differences to $[-\pi,\pi)$. Selection matrices $\mathbf{S}_1,\ldots,\mathbf{S}_4$ extract coordinates $\{1,2\}$, $\{3\}$, $\{4,5\}$, and $\{6\}$, corresponding to horizontal position, height, roll-pitch, and yaw errors. These errors supervise accumulated poses, whereas fusion operates on relative body-frame increments.

The sequence objective combines trajectory error and temporal regularization:
\begin{equation}
\begin{aligned}
    \mathcal{L}_{\tau}
    ={}&
    \frac{1}{K}
    \sum_{k=1}^{K}\sum_{\ell=1}^{4}
    \frac{e_{k,\ell}^{2}}
         {\operatorname{sg}(e_{k,\ell})+\epsilon}\\
    &+
    \frac{\lambda_{\alpha}}{K-1}
    \sum_{t=1}^{K-1}
    \left\|
        \boldsymbol{\alpha}_t
        -\boldsymbol{\alpha}_{t-1}
    \right\|_2^2.
\end{aligned}
\label{eq:gate_training}
\end{equation}
Here, $\epsilon>0$ prevents division by zero, and the detached denominator reduces the dependence of direct residual-gradient magnitudes on error magnitude. Position and angular errors use meters and radians without additional group weights. The coefficient $\lambda_{\alpha}\geq0$ controls the penalty on abrupt fusion changes.

Only $\theta$ is optimized. We detach query poses through $\mathcal{Q}(\mathcal{M};\operatorname{sg}(\hat{\mathbf{T}}_t))$ and proposal features entering the gate through Eq.~\eqref{eq:gate_context}. Specialist outputs used in fusion remain differentiable with respect to predicted-state inputs, allowing gradients through the fused trajectory and gate recurrence despite frozen specialist parameters.

\section{Implementation}
\label{sec:implementation}

We describe the data collection and preprocessing, terrain representation and soil prediction, recurrent gate design and training, and RL policy optimization using Verti-WM.

\subsection{Data Collection and Preprocessing}
\label{sec:implementation_data}

Simulation trajectories are collected in Verti-Bench~\cite{vertibench2025}, which uses Chrono~\cite{tasora2016chrono} to simulate vehicle-terrain interactions. The simulation data comprise rigid-terrain, deformable-terrain, and mixed-terrain collections, denoted by $\mathcal{D}_r$, $\mathcal{D}_d$, and $\mathcal{D}_m$, respectively. Each collection contains $20$ worlds with $25$ trajectories per world, yielding $500$ trajectories per collection. Each trajectory lasts approximately $25$~s. Together, the three collections contain $1{,}500$ trajectories and approximately $10.4$~hours of interaction data.

Randomized sinusoidal steering and speed commands generate diverse vehicle motions. Steering and speed frequencies are sampled from $[0.1,0.5]$~Hz and $[0.1,2.5]$~Hz, respectively, while the lower and upper speed bounds are sampled from $[1,2]$~m/s and $[3,4]$~m/s. Vehicle states, control inputs, and aligned elevation and semantic observations are recorded at $10$~Hz, giving a transition interval of $\Delta t=0.1$~s. This produces approximately $125{,}000$ transitions per collection and $375{,}000$ transitions overall.

Physical trajectories are collected on the Verti-Arena~\cite{chen2025verti} with the Verti-4-Wheeler~\cite{datar2024toward}, recording synchronized vehicle states, control inputs, and terrain observations. The physical dataset contains $120$ trajectories, each lasting approximately $35$~s, yielding approximately $70$~minutes of interaction data. These trajectories are used to train a separate Verti-WM from physical vehicle-terrain interactions.

\subsection{Terrain Representation and Soil Prediction}
\label{sec:implementation_terrain}

At each predicted pose, the terrain map provides aligned $128\times128$ elevation and RGB semantic patches covering approximately $12.8\times12.8$~m. Both modalities use nearest-neighbor sampling. Elevation values are globally normalized to $[-1,1]$, and RGB semantic values are normalized channel-wise to the same range. Identical preprocessing is applied during training and autoregressive rollout.

Elevation and semantic observations are encoded by separate U-Net autoencoders with one-channel and three-channel inputs, respectively. Each encoder produces an $18\times16\times16$ latent tensor. The autoencoders are pretrained independently using mean squared reconstruction error, after which the encoders are frozen. NeSAM and the Transformer retain architecturally identical encoder pairs pretrained on the deformable and rigid collections, respectively. Each pair contains $21{,}514$ encoder parameters. The corresponding complete autoencoder pair contains $54{,}646$ parameters. Both encoder pairs process the same terrain patches while preserving the pretrained feature representations expected by their corresponding specialists.

A single soil predictor processes NeSAM's semantic features through two convolutional layers with channel widths $18\rightarrow32\rightarrow64$ and a multilayer perceptron with widths $64\rightarrow64\rightarrow6$. Hidden layers use SiLU activations, and spatial aggregation yields one six-parameter soil estimate per crop, consumed only by NeSAM. The soil predictor contains $28{,}262$ parameters, counted separately from the encoder pairs. These parameters are also remain frozen during gate training.

\subsection{Recurrent Gate Architecture and Training}
\label{sec:implementation_gate}

The gate input $\mathbf{c}_t$ in Eq.~\eqref{eq:gate_context} contains $68$ dimensions: $36$ terrain features, $12$ vehicle-state components, $2$ control inputs, two six-dimensional specialist motion proposals, and their six-dimensional difference. A linear layer projects this input to $64$ dimensions, which are processed by a Gated Recurrent Unit (GRU) with a $64$-dimensional hidden state.

The updated hidden state is concatenated with the original $68$-dimensional input. The resulting $132$-dimensional vector passes through a $132\rightarrow32\rightarrow4$ multilayer perceptron with a SiLU hidden activation. Its four unconstrained outputs form $\mathbf{o}_t$, which is mapped to fusion coefficients using Eq.~\eqref{eq:gate_coefficients}. These coefficients control planar translation, vertical translation, roll and pitch, and yaw, and are expanded to the six pose components using Eq.~\eqref{eq:weight_expansion}. The gate contains $33{,}764$ trainable parameters.

After independently training the two specialists, we freeze their parameters, terrain encoders, and soil predictor, and optimize only the gate on $\mathcal{D}_m$. Training windows contain $K=16$ supervised transitions, with consecutive window starts separated by four transitions. Each minibatch contains $64$ windows. Specialist histories are initialized from the preceding recorded context, and the gate hidden state is reset to zero at the beginning of each window.

\subsection{Reinforcement Learning Policy Training}
\label{sec:implementation_rl}

We use the frozen Verti-WM as the simulator for downstream off-road navigation policy optimization. The RL environment runs $20$ parallel imagined episodes. At each step, the policy observes the vehicle state together with local elevation and semantic terrain features queried at the predicted pose, and outputs a two-dimensional command consisting of speed and steering. The reward encourages goal-directed and stable traversal by rewarding progress toward the goal while penalizing excessive roll and pitch, vertical motion, and vehicle stalling. Episodes terminate upon goal completion, departure from the valid map region, prediction of a physically implausible vehicle state, or the maximum episode horizon.

To evaluate whether Verti-WM supports different policy-learning paradigms, we train Proximal Policy Optimization (PPO), Soft Actor-Critic (SAC), and Twin Delayed Deep Deterministic Policy Gradient (TD3) within the same learned simulator. All three methods use identical observation and action spaces, reward and termination definitions, and $20$ parallel environments. PPO collects $300$ steps per environment per update, yielding $6{,}000$ imagined transitions per iteration, and all methods are trained for $2{,}000$ iterations using their respective on-policy or off-policy update procedures. Verti-WM remains frozen throughout training, while its $18\times16\times16$ terrain embeddings are further compressed to a $16$-dimensional representation by a trainable policy-side encoder optimized jointly with the policy. This common interface enables Verti-WM to support both on-policy and off-policy reinforcement learning methods.


\section{Experiments}
\label{sec:experiments}

We first examine the effects of model family and training terrain to motivate the selection of kinodynamic specialists. We then compare Verti-WM against unified Transformer models trained on rigid, deformable, and mixed terrain with comparable dynamics parameter counts. Finally, we evaluate policy-training runtime and navigation performance, with all policies tested in high-fidelity Chrono environments.

\subsection{Kinodynamic Specialist Ablation}
\label{sec:experiments_specialists}

We evaluate NeSAM~\cite{nesam} and Transformer models trained separately on $\mathcal{D}_r$, $\mathcal{D}_d$, and $\mathcal{D}_m$. All prediction experiments use held-out trajectories from the mixed-terrain collection in Verti-Bench~\cite{vertibench2025}, partitioned into rigid-terrain, deformable-terrain, and transition windows. Each model performs a $16$-step autoregressive rollout from the same initial state and recorded control sequence, querying terrain observations at its own predicted poses. We report position prediction error at the final step in meters. 

\begin{table}[h]
    \centering
    \caption{Specialist comparison using $16$-step position prediction error in meters on the mixed-terrain test set. Rigid-trained and deformable-trained models are reported on their corresponding terrain subsets, while mixed-trained models are evaluated across all subsets.}
    \label{tab:specialist_ablation}
    \small
    \setlength{\tabcolsep}{3pt}
    \renewcommand{\arraystretch}{1.12}
    \begin{tabular}{lcccc}
        \toprule
        Model & Training data & Rigid $\downarrow$ & Deformable $\downarrow$ & Transition $\downarrow$ \\
        \midrule
        NeSAM & $\mathcal{D}_r$ & 0.6158 & -- & -- \\
        Transformer & $\mathcal{D}_r$ & \textbf{0.4385} & -- & -- \\
        \midrule
        NeSAM & $\mathcal{D}_d$ & -- & \textbf{0.8081} & -- \\
        Transformer & $\mathcal{D}_d$ & -- & 0.9684 & -- \\
        \midrule
        NeSAM & $\mathcal{D}_m$ & 0.5630 & \textbf{0.7751} & \textbf{0.7371} \\
        Transformer & $\mathcal{D}_m$ & \textbf{0.5310} & 0.8890 & 0.8832 \\
        \bottomrule
    \end{tabular}
    \vspace{-2pt}
\end{table}

Table~\ref{tab:specialist_ablation} shows complementary strengths under terrain-specific training. When trained on $\mathcal{D}_r$, the Transformer achieves a rigid-terrain error of $0.4385$~m, compared with $0.6158$~m for NeSAM. When trained on $\mathcal{D}_d$, NeSAM achieves a deformable-terrain error of $0.8081$~m, compared with $0.9684$~m for the Transformer. These comparisons motivate using the rigid-trained Transformer and deformable-trained NeSAM as the two specialists in Verti-WM.

The models trained directly on $\mathcal{D}_m$ exhibit a similar division of strengths: the Transformer provides lower rigid-terrain error, while NeSAM provides lower deformable-terrain and transition errors. Meanwhile, the rigid-trained Transformer incurs substantially higher error on deformable terrain, and the deformable-trained NeSAM remains less accurate on rigid terrain than the rigid-trained Transformer. These results motivate combining their predictions according to the evolving vehicle-terrain interaction.

\subsection{Comparison with Full-Data World Models}
\label{sec:experiments_prediction}

We next evaluate whether recurrent fusion improves prediction relative to unified models with comparable dynamics capacity. Our primary baselines are two Transformer models trained on $\mathcal{D}_{\mathrm{all}}=\mathcal{D}_r\cup\mathcal{D}_d\cup\mathcal{D}_m$. Both receive vehicle-state and control histories, while the exteroceptive variant additionally receives elevation and semantic terrain features. Their dynamics networks contain approximately $9.9$ million parameters, closely matching the $9{,}972{,}634$ parameters in Verti-WM's two specialists and recurrent gate.

Verti-WM uses the specialists selected in Sec.~\ref{sec:experiments_specialists}, freezes their parameters, and trains only the $33{,}764$-parameter gate on $\mathcal{D}_m$. Thus, all three components contribute to constructing Verti-WM, but only the gate is updated during mixed-terrain training. Both full-data Transformer baselines update their complete dynamics networks using all three collections. All models follow the same $16$-step prediction protocol.

The Transformer baselines without and with terrain observations contain $9{,}888{,}024$ and $9{,}871{,}128$ dynamics parameters, respectively, compared with $9{,}972{,}634$ for Verti-WM. These counts exclude terrain encoders and the separate soil predictor. Both baselines train their complete dynamics networks on $\mathcal{D}_{\mathrm{all}}$, whereas Verti-WM updates only its $33{,}764$ gate parameters on $\mathcal{D}_m$ after freezing the specialists pretrained on $\mathcal{D}_r$ and $\mathcal{D}_d$.

\begin{table}[h]
    \centering
    \caption{Comparison with full-data Transformer world models using $16$-step position prediction error in meters. Extero. indicates exteroceptive terrain conditioning. For Verti-WM, the training-data column refers to gate training after specialist pretraining on $\mathcal{D}_r$ and $\mathcal{D}_d$.}
    \label{tab:wm_prediction}
    \small
    \setlength{\tabcolsep}{4pt}
    \renewcommand{\arraystretch}{1.12}
    \begin{tabular*}{\columnwidth}{@{}l@{\extracolsep{\fill}}cccc@{}}
        \toprule
        Model
        & Training data
        & Rigid $\downarrow$
        & Deformable $\downarrow$
        & Transition $\downarrow$ \\
        \midrule
        \makecell[l]{Transformer \\ w/o extero.}
        & $\mathcal{D}_{\mathrm{all}}$
        & 0.4674 & 0.7905 & 0.7522 \\
        \addlinespace[2pt]
        \makecell[l]{Transformer \\ w/ extero.}
        & $\mathcal{D}_{\mathrm{all}}$
        & \textbf{0.3413} & 0.6964 & 0.6518 \\
        \addlinespace[2pt]
        Verti-WM
        & $\mathcal{D}_m$ (gate)
        & 0.3655 & \textbf{0.6409} & \textbf{0.5773} \\
        \bottomrule
    \end{tabular*}
    \vspace{-2pt}
\end{table}

Table~\ref{tab:wm_prediction} shows that incorporating terrain observations improves the full-data Transformer across all three evaluation subsets. Relative to the Transformer without terrain observations, Verti-WM reduces prediction error by $21.8\%$, $18.9\%$, and $23.3\%$ on rigid, deformable, and transition windows, respectively.

\begin{table*}[!ht]
    \centering
    \caption{Policy-training runtime and navigation performance in Chrono. Training time records each run's total duration, excluding dataset collection and world-model training. Traversal time, roll, and pitch are averaged over successful trials.}
    \label{tab:rl_results}
    \small
    \setlength{\tabcolsep}{6pt}
    \renewcommand{\arraystretch}{1.12}
    \begin{tabular}{lccccccc}
        \toprule
        Metric
        & \shortstack{Chrono\\PPO}
        & \shortstack{Transformer\\PPO}
        & \shortstack{NeSAM\\PPO}
        & \shortstack{Verti-WM\\PPO}
        & \shortstack{Verti-WM\\TD3}
        & \shortstack{Verti-WM\\SAC}
        & \shortstack{Verti-WM\\PPO (CPU)} \\
        \midrule
        Training time (h)
        & 33.48
        & 1.14
        & \textbf{1.01}
        & 1.36
        & 1.35
        & 1.55
        & 3.98 \\

        Success count $\uparrow$
        & 18/20
        & 17/20
        & 17/20
        & 19/20
        & 18/20
        & 18/20
        & \textbf{20/20} \\

        Traversal time (s) $\downarrow$
        & \textbf{27.5 $\pm$ 4.2}
        & 28.1 $\pm$ 3.9
        & 29.7 $\pm$ 7.0
        & \textbf{27.5 $\pm$ 3.0}
        & 32.2 $\pm$ 5.1
        & 31.3 $\pm$ 7.5
        & 34.0 $\pm$ 11.6 \\

        $|\mathrm{Roll}|$ ($^\circ$) $\downarrow$
        & \textbf{4.37 $\pm$ 1.88}
        & 4.85 $\pm$ 2.27
        & 4.42 $\pm$ 1.85
        & 4.84 $\pm$ 2.21
        & 4.64 $\pm$ 1.90
        & 4.88 $\pm$ 2.23
        & 4.88 $\pm$ 2.23 \\

        $|\mathrm{Pitch}|$ ($^\circ$) $\downarrow$
        & 3.69 $\pm$ 1.30
        & 4.06 $\pm$ 2.01
        & \textbf{3.62 $\pm$ 1.38}
        & 3.76 $\pm$ 1.35
        & 3.65 $\pm$ 1.30
        & 3.86 $\pm$ 1.56
        & 3.93 $\pm$ 1.78 \\
        \bottomrule
    \end{tabular}
    \vspace{-5pt}
\end{table*}

Compared with the full-data exteroceptive Transformer, Verti-WM reduces error by $8.0\%$ on deformable terrain and $11.4\%$ across terrain transitions. The exteroceptive Transformer achieves the lowest rigid-terrain error, with $0.3413$~m compared with $0.3655$~m for Verti-WM. The benefit of recurrent specialist fusion is therefore concentrated on deformation-dependent motion and transitions between terrain regimes.

Verti-WM also improves over both constituent specialists and both models trained directly on $\mathcal{D}_m$ in Table~\ref{tab:specialist_ablation} across all three subsets. Together, these results show that recurrent fusion can improve mixed-terrain prediction while updating only $0.34\%$ of the combined dynamics parameters during gate training.

\subsection{Policy Training and Evaluation in Verti-Bench}
\label{sec:experiments_rl}

We compare policy training in Chrono with training inside three frozen learned environments: the rigid-terrain Transformer, the deformable-terrain NeSAM model, and Verti-WM. PPO provides the primary comparison across training environments. Additional runs use TD3 and SAC within Verti-WM. We also evaluate PPO training with Verti-WM's dynamics executed on a CPU.

All policies are subsequently evaluated in the same Chrono environments in Verti-Bench~\cite{vertibench2025}. Evaluation covers $20$ worlds, and success is reported as the number of completed trials. Traversal time is averaged over successful trials only, whereas absolute roll and pitch are averaged over all recorded steps, including failed trials. Training time denotes the recorded duration of each policy-training run, excluding dataset collection and world-model training.

Table~\ref{tab:rl_results} shows that the PPO policy trained within Verti-WM completes $19$ of $20$ trials in Chrono, compared with $18$ for direct Chrono training and $17$ for training within either individual specialist. Its mean traversal time is $27.5$~s, matching the direct-training baseline, with mean absolute roll and pitch of $5.02^\circ$ and $3.94^\circ$, respectively. These results demonstrate that policies optimized entirely within Verti-WM can support off-road navigation after transfer to high-fidelity dynamics.

Direct Chrono training takes $33.48$~hours, whereas the recorded Verti-WM PPO run takes $1.36$~hours. When Verti-WM's dynamics are executed on the CPU, policy training takes $3.98$~hours, corresponding to an $8.41\times$ ratio in recorded runtime and an $88.1\%$ reduction relative to direct Chrono training under the reported configurations. The CPU-trained policy completes all $20$ trials, with mean absolute roll and pitch of $4.84^\circ$ and $3.82^\circ$. Its mean traversal time of $34.0$~s covers all $20$ successful trials, while the direct-training baseline averages its $18$ successful trials.

Verti-WM also supports policy training with TD3 and SAC, whose policies each complete $18$ of $20$ trials in Chrono after recorded training runtimes of $1.35$ and $1.55$~hours, respectively. Their mean traversal times are $32.2$ and $31.3$~s, compared with $27.5$~s for PPO trained within Verti-WM. These results demonstrate that the learned environment supports multiple RL algorithms while enabling policy deployment under high-fidelity vehicle-terrain dynamics.

\subsection{Physical-Vehicle Evaluation on Verti-Arena}
\label{sec:exp_physical}

\begin{figure}[!t]
    \centering
    \includegraphics[width=\columnwidth]{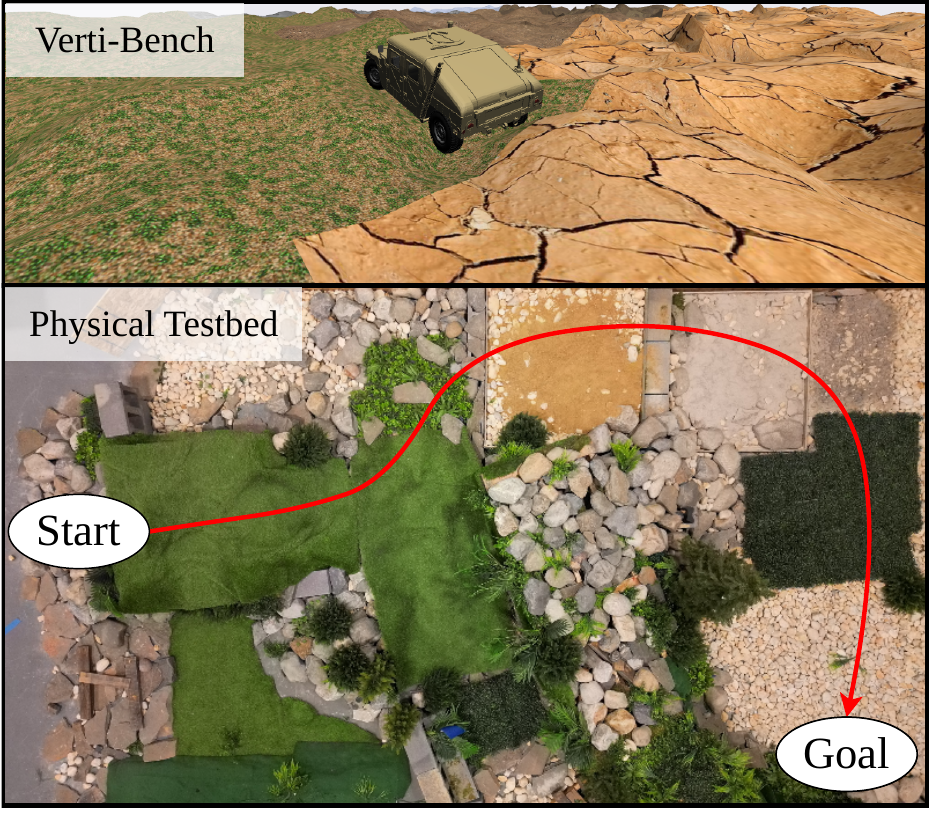}
    \caption{Evaluation environments in Verti-Bench (top) and physical testbed (bottom). The red arrow indicates the navigation route from start to goal.}
    \label{fig:evaluation_environments}
    \vspace{-10pt}
\end{figure}

We evaluate whether learning a world model from physical interactions improves policy deployment relative to direct sim-to-real transfer. We construct a separate Verti-WM from the physical trajectories described in Sec.~\ref{sec:implementation_data} and optimize a navigation policy within this model. The baseline policy is trained in the learned simulation from physical data and transferred directly to the Verti-4-Wheeler without fine-tuning. Each policy is evaluated in five physical trials in the Verti-Arena environment shown in Fig.~\ref{fig:evaluation_environments}, navigating between the marked start and goal.

\begin{table}[h]
    \centering
    \caption{Physical navigation performance on the Verti-4-Wheeler. All metrics except success are reported as mean $\pm$ standard deviation over successful trials.}
    \label{tab:physical_results}
    \small
    \setlength{\tabcolsep}{4pt}
    \renewcommand{\arraystretch}{1.12}
    \begin{tabular}{lcc}
        \toprule
        Metric & \shortstack{Direct sim2real } & \shortstack{Physical Verti-WM} \\
        \midrule
        Success count $\uparrow$ & 2/5 & \textbf{4/5} \\
        Traversal time (s) & $\textbf{33.44} \pm \textbf{4.88}$ & $39.53 \pm 4.77$ \\
        $|\mathrm{Roll}|$ ($^\circ$) & $7.53 \pm 1.42$ & $\textbf{6.16} \pm \textbf{0.52}$ \\
        $|\mathrm{Pitch}|$ ($^\circ$) & $11.26 \pm 1.16$ & $\textbf{9.37} \pm \textbf{0.50}$\\
        \bottomrule
    \end{tabular}
    \vspace{-10pt}
\end{table}

As shown in Table~\ref{tab:physical_results}, the policy optimized within the real-data Verti-WM completes four of five physical trials, compared with two of five for direct sim-to-real transfer. Among successful trials, it achieves lower mean absolute roll and pitch, compared with direct transfer. Its mean traversal time is longer, at $39.53$~s versus $33.44$~s. The higher observed completion rate supports constructing a learned environment from physical vehicle-terrain interactions for policy optimization, while the attitude improvements characterize successful traversals only.

\section{Conclusion}
\label{sec:conclusion}

We presented \textit{Verti-WM}, a physics-aided exteroceptive world model that recurrently fuses frozen kinodynamic specialists to generate map-conditioned 6-DoF rollouts for off-road RL. Experiments show improved prediction on deformable terrain and terrain transitions, together with comparable navigation success and shorter policy-training runtimes than direct high-fidelity simulator training. A policy optimized within a separate model learned from physical trajectories completes four of five trials, compared with two of five for direct sim-to-real transfer.

The current formulation relies on a ground truth elevation and semantic map. Future work will investigate online map updates and uncertainty-aware rollouts to support policy learning under incomplete terrain observations.



\bibliographystyle{IEEEtran}
\bibliography{IEEEabrv,references}

\end{document}